\documentclass[conference]{IEEEtran}
\usepackage{graphicx} 
\usepackage{amsmath, amssymb}
\def\BibTeX{{\rm B\kern-.05em{\sc i\kern-.025em b}\kern-.08em
    T\kern-.1667em\lower.7ex\hbox{E}\kern-.125emX}}

\def\real{\mathbb{R}}
\def\trans{^\mathrm{T}}
\def\Ik{s_k}
\newcommand{\Arow}[1]{A_{#1}}
\def\basephases{\boldsymbol{\alpha}}
\newcommand{\Acol}[1]{\basephases_{#1}}
\newcommand{\Crow}[1]{C_{#1}}
\newcommand{\CArow}[1]{\Crow{#1}A}

\def\corrupt{{\bf u}}
\def\clean{{\bf v}}

\def\symbolA{{\bf T}}
\def\symbolB{{\bf S}}
\def\objectA{triangle}
\def\objectB{square}
\def\coloursymbolA{{\bf R}}
\def\colourA{red}
\def\X{{\bf X}}
\def\Y{{\bf Y}}
\def\M{{\bf M}}

\def\myphi{\boldsymbol{\phi}}
\def\myk{\boldsymbol{k}}
\def\mytheta{\boldsymbol{\theta}}

\def\Cu{\mathbb{C}_1}

\DeclareMathOperator*{\argmin}{argmin}
\DeclareMathOperator*{\argmax}{argmax}
\newcommand{\cip}[2]{\left< #1 , #2 \right>}
\newcommand{\grad}[1]{\nabla_{\hspace{-0.8mm}#1}}

\title{Improved Cleanup and Decoding of Fractional Power Encodings}
\author{Alicia Bremer, Jeff Orchard\\
Cheriton School of Computer Science, University of Waterloo}
\date{August 2024}

\begin{document}

\maketitle

\begin{abstract}
High-dimensional vectors have been proposed as a neural method for representing information in the brain using Vector Symbolic Algebras (VSAs). While previous work has explored decoding and cleaning up these vectors under the noise that arises during computation, existing methods are limited. Cleanup methods are essential for robust computation within a VSA. However, cleanup methods for continuous-value encodings are not as effective. In this paper, we present an iterative optimization method to decode and clean up Fourier Holographic Reduced Representation (FHRR) vectors that are encoding continuous values. We combine composite likelihood estimation (CLE) and maximum likelihood estimation (MLE) to ensure convergence to the global optimum. We also demonstrate that this method can effectively decode FHRR vectors under different noise conditions, and show that it outperforms existing methods.
\end{abstract}

\section{Introduction}

Vector Symbolic Algebras (VSAs), also known as hyperdimensional computing, offer a way to represent symbols as high-dimensional vectors. These symbol vectors can then be combined to form compound symbols, which can further be combined with other symbols, or decomposed back into their constituent vectors. For example, we could have a vector $\symbolA$ that represents a ``\objectA'', and another vector $\coloursymbolA$ that represents ``\colourA'', and combine those two into another vector that represents ``a \colourA \ \objectA''. In this way, VSAs enable a compositional language within the substrate of a single vector space, and have been proposed as a way to perform computations in the brain \cite{smolensky1990,kanerva2009,eliasmith2013}, including a solution to the binding problem \cite{Gayler2003}.

The basic functions of a VSA are carried out by a small set of vector operations: similarity, binding, bundling, and clean-up. The \emph{similarity} operation, $\cdot$, takes two vectors and computes how similar they are, yielding a value of 1 if the vectors are identical, and 0 if they are completely different. We would expect the value of $\coloursymbolA \cdot \symbolA$ to be close to zero. The \emph{binding} operation, $\otimes$, combines two vectors and creates a composite vector that is dissimilar to each of its inputs. For example, the concept of ``a \colourA \ \objectA'' is represented by the vector $\coloursymbolA \otimes \symbolA$. Importantly, binding is invertible (using the operator $\oslash$) so that the resulting vector can be factored back into its composite vectors. The operation of \emph{bundling}, $\oplus$, also combines vectors; but unlike binding, bundling combines vectors in a way that the resulting vector is still similar to its component vectors. In our example, the vector $\coloursymbolA \oplus \symbolA$ is a vector that represents the collection concepts of ``\colourA'' and ``\objectA'', but not ``a \colourA \ \objectA''.

Hence, given a vocabulary of concepts and their corresponding vectors, many different propositions can be constructed, manipulated, and queried. For example, we could construct the vector $\clean = \coloursymbolA \otimes \symbolA$, and then query it by asking about the properties of the \objectA \ using $\clean \oslash \symbolA$, which would result in a vector similar to $\coloursymbolA$. These VSA operations can introduce noise, which can reduce the fidelity of subsequent operations. For this reason, many VSA implementations have a \emph{clean-up} operation that takes a noisy or corrupted vector and restores it to the nearest recognized symbol (vector) in the vocabulary.

Some VSAs also have a facility to encode continuous values, not just discrete symbols. For example, we can encode the value 1.5 into the vector $\X$. Such vectors are often called Spatial Semantic Pointers, or SSPs, since these vectors can encode spatial location \cite{Komer2019}. Then the composite vector $\corrupt = \clean \otimes \X$ could represent ``a red \objectA \ at location 1.5''. That vector could be queried, asking where the \colourA \ \objectA \ is using $\corrupt \oslash \clean$. Ideally, the resulting vector would be $\X$. But in practice, the vector is corrupted by the sequence of VSA operations. Unfortunately, there is currently no clean-up mechanism for SSPs, since the uncorrupted vectors occupy a manifold in the vector space, not just a single point.

We could perform clean-up if we could robustly decode (and re-encode) the value encoded in an SSP. However, even though the algorithm to \emph{encode} a value into an SSP is relatively straight-forward, the inverse is not so simple, and there is currently no closed-form formula for \emph{decoding}.

Some attempts have been made to solve the decoding/clean-up problem for SSPs. Some methods perform a \emph{grid search}, comparing the noisy SSP to a large number of clean SSPs via the similarity operation, and picking the clean hypervector that is the most similar \cite{lu2019}. Another approach is to train a neural network as a \emph{denoising autoencoder} to do the clean-up \cite{komer2020,voelker2021a}; such networks have also been demonstrated using spiking neurons \cite{Stewart2011}.

A related problem is how to parse a composite hypervector into its constituent parts.
This is a challenging task, since each binding affects the vector, and decoding any one vector depends on knowing the composition of the remaining components. A \emph{resonator network} starts with a guess for each of the components, uses those guesses to try to factor out each component individually, performs a grid search for each component, and then reconstitutes the vector based on the best fit for each component \cite{frady2020, kleyko2022}. Least squares or gradient-based optimization methods can also be used to minimize a loss function between the given vector and the estimates for the components, but often converge to the wrong value when compared to resonator networks \cite{kent2020}.

In this paper, we propose a new optimization method for cleaning up and decoding SSP vectors, specifically within a type of VSA called a Fourier Holographic Reduced Representation. Since a value is encoded in an SSP using phase angles (described later), we optimize similarity using
circular distance regression methods \cite{lund1999}. We avoid convergence to local optima by also considering the coupling relationships between the phase angles. The result is an efficient and robust SSP decoding and clean-up method.

\section{Methods}

The goal is to find a clean SSP that maximizes its similarity to the noisy SSP. However, the similarity function often has many local optima, so it can be difficult to ensure convergence to the global optimum. Is there some way to rule out sub-optimal solutions?

Our idea is to regularize the optimization process using the relationships between the vector components. These couplings can guide the optimization process toward a global optimum.

This approach was inspired by oscillatory-inference (OI) models for path integration. In path integration, an animal estimates its location by integrating its own velocity. Hypothesized velocity-controlled oscillators (VCOs) vary their frequency according to the animal's velocity \cite{burgess2008}, and thereby encode position as phase differences. Any error in the oscillations will cause their phases to drift, and reduce the fidelity of the encoding. However, phase-coupling mechanisms can help reduce the drift \cite{OrchardFrontCompNeuro13,burgess2014, Ji2014}. These OI models of path integration are compatible with the SSP encoding of space, since each VCO can be thought of as an element of an SSP vector \cite{dumont2022model}.

\subsection{Vector Operations in FHRR}

We will focus on a VSA called the Fourier Holographic Reduced Representation (FHRR) \cite{Plate1995}; the FHRR is essentially the Fourier transform of another VSA called the Holographic Reduced Representation (HRR). These VSAs can efficiently store large sets of symbols, comparing favourably to other VSAs \cite{schlegel2022}. They can also produce SSPs using Fractional Power Encoding (FPE) \cite{plate1992, plate1994}, or fractional binding \cite{Komer2019}.

The entries of FHRR vectors are complex numbers. We specifically focus on unitary vectors, where the modulus of each complex number is $1$. We will refer to the set of unit-modulus complex numbers a $\Cu$. Thus, each element of $\Cu$ can be fully specified by a phase angle in the range $(-\pi,\pi]$. Given the vector of phases $\basephases=[a_1, \ldots , a_n]\trans$, we can write the $k$th element as $e^{i a_k}$, or the whole vector as $e^{i\basephases}$, where the exponent is applied element-wise to $\basephases$. We will refer to the vector of phases ($\basephases$, in this case) as the {\it base phases}.

The similarity operation for the FHRR is the complex inner product. For $\corrupt,\clean \in \Cu^n$,
$$
\left< \corrupt , \clean \right> = \sum_{k=1}^n \corrupt_k \, \overline{\clean}_k \ ,
$$
where $\overline{\clean}_k$ is the complex conjugate of $\clean_k$. For unitary vectors, the inner product is equivalent to computing cosine similarity.

The \emph{binding} operation, $\otimes$, for FHRR vectors in $\Cu^n$ is the Hadamard product, or element-wise multiplication, which can be written $e^{i\basephases} \otimes e^{i\boldsymbol{\beta}} = e^{i( \basephases + \boldsymbol{\beta})}$.

The \emph{bundling} operation, $\oplus$, for FHRR is simple vector addition. It is worth noting that such a bundle is usually not unitary, so has to be normalized to convert it back into a unitary vector. Bundles of SSPs have been used to represent functions, including probability densities \cite{furlong2022fractional}.

The \emph{FPE} or \emph{fractional binding} operation is a generalization of binding, where an FHRR vector can be bound to itself. Since binding is element-wise multiplication, we can bind an FHRR vector to itself using 
\begin{equation}
    \underbrace{ e^{i\basephases} \otimes e^{i\basephases} \otimes \cdots \otimes e^{i\basephases} }_{m \text{ times}} =  e^{i\basephases m} \ ,
\end{equation}
where $m \in \mathbb{Z}$. In FPE, the value of $m$ can be any real value, so we can encode $x \in \real$ using the SSP $e^{i\basephases x}$.

Representing a \underline{vector} of continuous values $\vec{x} = [x_1,\ldots, x_d]^\mathrm{T}$ is accomplished by encoding each element into an SSP (each using a different vector of $n$ base phases) and then binding all those SSPs into a single SSP. If we use $\Acol{k}$ to represent the vector of base phases for encoding $x_k$, then the multi-dimensional SSP is $e^{i \Acol{1}  x_1} \otimes \cdots \otimes e^{i \Acol{d} x_d} = e^{iA \vec{x}}$, where $A$ is a matrix with columns $\Acol{k}$, $k=1, \ldots, d$.

\subsection{Corruption of SSP Vectors}

Spatial semantic pointers can be corrupted, especially when bundling vectors. The fact that hypervectors are only \emph{approximately} orthogonal means that they will interfere with each other in a bundle. For example, when a bundle is queried using the unbinding operation, the lack of perfect orthogonality means that each vector in the bundle leaves a small residual.

Normalizing a bundle introduces even more corruption. Suppose we have a bundle composed of various objects, each bound with an SSP that encodes its location. This bundle represents a spatial map that can be used for navigation \cite{dumont2023b}. We can query that bundle either by location vector or by object vector \cite{lu2019, Komer2019}. As an example, consider the normalized bundle
\begin{equation}
    \M =  \symbolA \otimes \X^{1.5} \otimes \Y^{-2.3} + \symbolB \otimes \X^{-0.7} \otimes \Y^{0.3}
\end{equation}
where $\symbolA$ and $\symbolB$ are FHRR vectors representing ``\objectA'' and ``\objectB'', and $\X$ and $\Y$ are the SSPs representing the $x$- and $y$-axis. The bundle $\M$ represents a set with a \objectA \ at $(1.5, -2.3)$ and a \objectB \ at $(-0.7,0.3).$

We can query for the location of the \objectA \ using the unbinding operation, $\M \oslash \symbolA$, yielding a noisy SSP approximately representing the location $(1.5, -2.3)$.

\begin{figure}[!tb]
    \centering
    \includegraphics[width=0.4\textwidth]{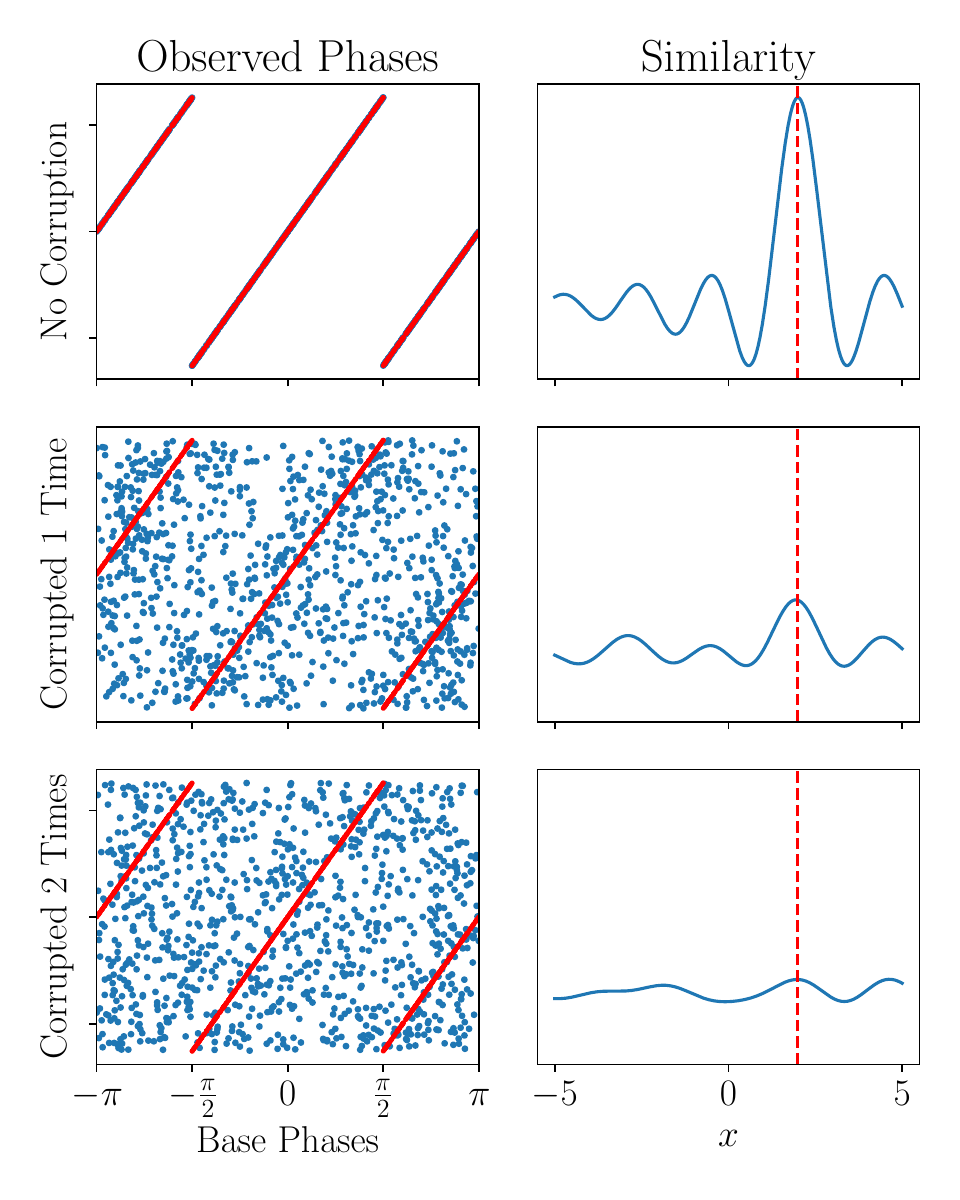}
    \caption{The uncorrupted (top), corrupted (middle), and twice corrupted (bottom) SSP phases (left) and similarity for different encoded $x$ values (right). The left plot displays the ideal phases in red and observed phases in blue. The right plot displays the similarity (blue) to SSPs encoding other $x$ values, and the true $x$-value (red).}
    \label{figure:corruption_example}
\end{figure}

If this process is repeated, that is, the noisy SSP is encoded into another bundle and queried again, the resulting SSP may no longer represent the location $(1.5, -2.3)$. The left-hand column of Fig.\,\ref{figure:corruption_example} shows how the phases get corrupted after repeated bundling/querying episodes, where the bundles included 13 vectors, all added together and then normalized. The right-hand column plots the similarity of the corrupted SSPs to uncorrupted SSPs across a range of $x$-values. After two bundling/querying steps, the correct $x$-value can no longer be determined.

From the bundle $\M$, we could also query for the object at location $(1.5, -2.3)$ by unbinding, $\M \oslash (\X^{1.5} \otimes \Y^{-2.3})$.
The result is a noisy version of the $\symbolA$ vector. If the SSP used to query the bundle is clean, the resulting vector should be most similar to the vector $\symbolA$. However, if the SSP was corrupted with noise, that noise will be carried through the unbinding operation; the resulting vector might be misidentified.

Noise can also arise due to imprecise computations with FHRR vectors. The computations performed during neural implementations of algorithms using FHRR or HRR vectors often introduce noise. One example is path integration, where the integration of the velocity input introduces noise \cite{dumont2022model, dumont2023b}.

In all these examples, having a clean-up operation for SSPs is important, and we demonstrate the utility of our method on some of these examples in Section \ref{section:experiments}.



\subsection{Least Circular Distance Regression on SSP Phases}\label{direct}

\begin{figure}
    \includegraphics[width=0.48\textwidth]{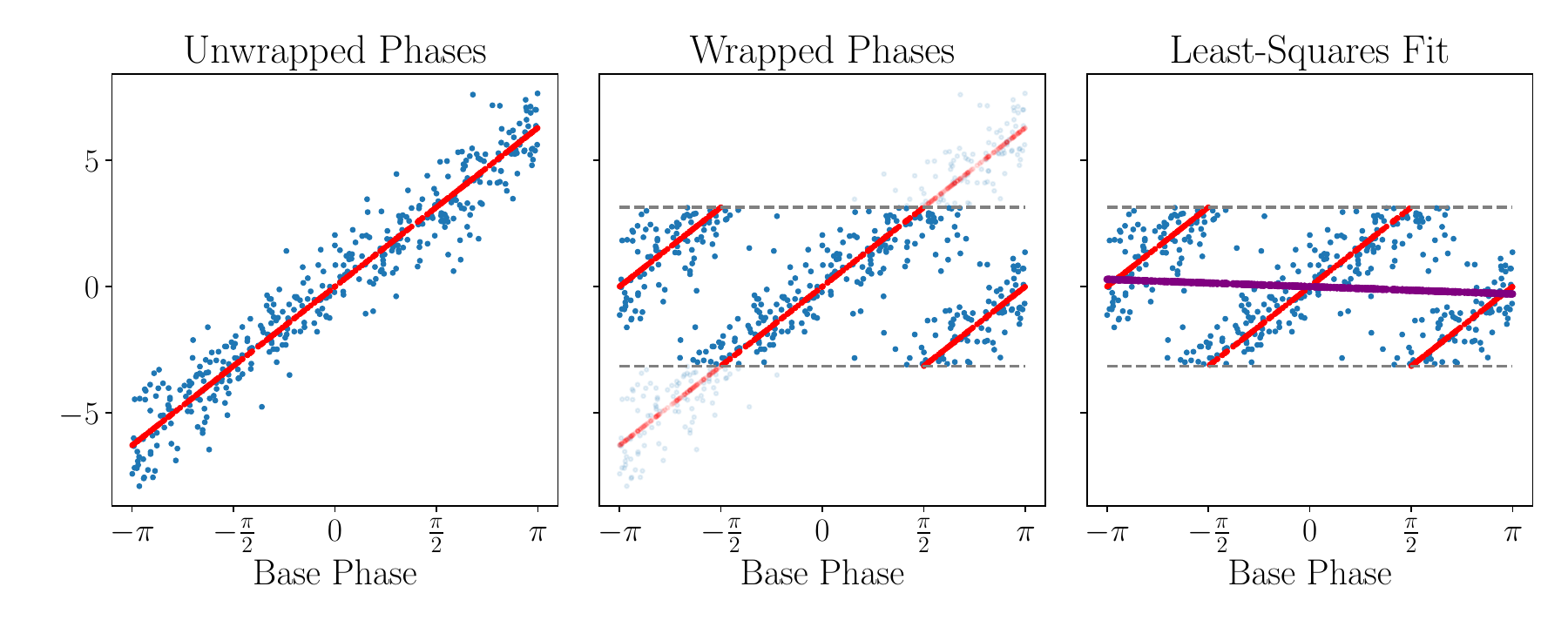}
    \caption{Failure of linear least-squares. For $x=2$, the left plot shows the unwrapped phases (blue dots) and the linear model $Ax$ (red line). The middle plot shows the same, but with wrapping to $(-\pi,\pi]$. The plot on the right shows the linear least-squares fit (purple) to the wrapped phases. That fit is not close to the true model (red).}
    \label{figure:phase_visualization}
\end{figure}

Suppose we have an SSP vector, $\corrupt$, encoding $\vec{x}$. Since $\corrupt$ is unitary, we can write it as
\begin{equation}
\label{eq:ssp}
    \corrupt = e^{i\myphi} \approx e^{i A \vec{x}}
\end{equation}
where $\myphi$ is the vector of phases observed in $\corrupt$. The task of decoding $\corrupt$ is to find $\vec{x}$ that maximizes the real part of the similarity between the complex vectors $\corrupt$ and $e^{iA\vec{x}}$,
\begin{equation}
    S(\vec{x}) = \mathrm{Re} \left( \corrupt \cdot \overline{e^{iA\vec{x}}} \right) = \mathrm{Re} \left( e^{i (\myphi - A \vec{x})} \right) . \label{eq:real_sim}
\end{equation}
The similarity would be maximized if $\myphi$ equaled $A\vec{x}$.
So it might seem sensible to try to choose $\vec{x}$ to minimize the phase error as a least-squares problem,
$$
\argmin_{\vec{x}} \| \myphi - A\vec{x} \|_2^2 \ .
$$
However, the complex exponential function is periodic, so that a phase of $-\pi$ is the same as a phase of $\pi$. The phases we extract from $\corrupt$ will be wrapped to the range $(-\pi, \pi]$, while the phases $A\vec{x}$ are not wrapped. Unfortunately, phase wrapping makes it difficult to compare $\myphi$ and $A\vec{x}$ directly. Figure\,\ref{figure:phase_visualization} demonstrates how linear least-squares regression fails due to phase wrapping.

We could unwrap $\corrupt$ by adding or subtracting multiples of $2\pi$, and then try to solve the unwrapped least-squares problem,
\begin{equation} \label{eq:linear_ls}
\argmin_{\vec{x},\myk\in \mathbb{Z}^n} \| \myphi  + 2\pi \myk - A\vec{x} \|^2 \ .
\end{equation}
But that requires we jointly solve an integer optimization problem to find $\myk$.

Instead, we can formulate (\ref{eq:linear_ls}) as a least \emph{circular} distance (LCD) problem because the base phases and expected phases are both angular measurements \cite{lund1999}.
\begin{equation}
\label{eq:lcd}
    \argmin_{\vec{x}} \frac1{2n} \sum_{i=1}^n \Big( 1 - \cos(\phi_i -\Arow{i} \vec{x} ) \Big)
\end{equation}
%
Since the cosine function is $2\pi$-periodic, this approach allows us to compare $\myphi$ and $A\vec{x}$ directly without having to worry about wrapping. We can rewrite it as $\argmax_{\vec{x}} E_D(\vec{x})$, where
\begin{equation}
\label{eq:lcd_simplfied}
E_D(\vec{x}) = \frac1n \sum_{i=1}^n  \cos(\phi_i -\Arow{i} \vec{x} ) \ .
\end{equation}
It turns out that (\ref{eq:real_sim}) is the same as (\ref{eq:lcd_simplfied}), so that $S(\vec{x}) = n E_D(\vec{x})$.

If each $\phi_i$ is the observed value of a circular random variable $\Phi_i$ that follows a von Mises distribution (an approximation of the normal distribution for circular data) centred at $\Arow{i} \vec{x}$ and with variance $\kappa$, then the maximum likelihood estimate (MLE) of $\vec{x}$ given $\phi_1,\ldots, \phi_n$ maximizes \eqref{eq:lcd_simplfied} \cite{lund1999}.

The connection between the MLE and optimal decoding from grid cell populations representing location has been previously explored \cite{Sreenivasan2011}. In that work, the phases were modelled by a slightly different circular distribution. Discrete $\vec{x}$ values were considered, making that method more similar to a grid search over all possible $\vec{x}$ values. 


The similarity operation has been previously explored as a probability measure for fractional binding in VSAs. With modifications to the similarity operation to ensure non-negative values, computing the similarity between $\corrupt$ and $e^{iA \vec{x}}$ can give the probability of $\vec{x}$ \cite{furlong2022fractional}.

Solving the optimization problem, \eqref{eq:lcd} or \eqref{eq:lcd_simplfied}, can decode $\vec{x}$ from an SSP. Then, clean-up is achieved by re-encoding this decoded value as a clean SSP. 

However, the objective function in \eqref{eq:lcd_simplfied} will often have many local minima and maxima (for example, see the top-right plot in Fig.\,\ref{figure:corruption_example}). This non-convex function is challenging to maximize, since iterative optimization methods often fall into the wrong basin of attraction and converge to a suboptimal solution. In the next section, we propose a slightly different objective function that is far less susceptible to this local optimum problem.



\subsection{Phase-Coupled Least Circular Distance Regression}\label{couplings}

The local optima in the similarity function occur because, as $\vec{x}$ diverges from its optimal value, the differences between $\phi_i$ and $\Arow{i} \vec{x}$ diverge at first, but then converge again as $\Arow{i} \vec{x}$ completes a full lap and returns close to $\phi_i$. This problem is made worse by large base phases since bigger $\Arow{i}$ results in faster-growing phases. However, the \emph{differences} between these phases need not be large. By choosing two base phases that are closer together, we can expect their phase difference to be smaller and less prone to wrapping. 


A similar type of phase coupling was used to correct phase errors for path integration \cite{Ji2014}. A web of phase couplings was compiled into an over-determined linear system, the least-squares solution of which yielded the cleaned-up phases as well as an estimate of $\vec{x}$. This formulation reduced the impact of the phase wrapping problem. We can use the idea of strategically coupling phases to formulate a least circular distance problem with a smoother, less undulating objective function.

Consider vector elements $i$ and $j$, with observed phases $\phi_i$ and $\phi_j$. Their phase difference can be written
\begin{align}
    \Delta \phi_{i,j} & = \phi_i - \phi_j \\
    & \approx \Arow{i} \vec{x} - \Arow{j} \vec{x} \\
    & = (\Arow{i}-\Arow{j}) \vec{x} \\
    & = \Delta A_{i,j} \vec{x} \ .
\end{align}
Similarly, their phase sum is
\begin{equation}
    \overline{\Delta} \phi_{i,j} = \phi_i + \phi_j \approx \overline{\Delta} A_{i,j} \vec{x} \ .
\end{equation}

\def\ncouplings{n_c}

We can construct $\ncouplings$ pairs of coupled phases, and thus define an $\ncouplings \times n$ matrix $C$, which we will call the \emph{coupling matrix}. If the $k$th coupling is between phases $\phi_{i_k}$ and $\phi_{j_k}$, then we set the $k$th row of $C$, denoted $\Crow{k}$, to zero, except for $C_{k,i_k}=1$ and $C_{k, j_k} = \Ik$. Here $\Ik$ is 1 if we sum the phases, and -1 if we subtract the phases. Whether we are adding or subtracting phases, we will refer to these phase combinations as \emph{phase differences}.

From this definition, $C\myphi$ is a vector of the phase differences for all the couplings. Similarly, $C A$ is a vector of base phase differences for all the couplings. Then, the least circular distance formulation is 
\begin{align}
    \argmin_{\vec{x}} \frac1{\ncouplings} \sum_{k=1}^{\ncouplings} \Big( 1 - \cos \left( \Crow{k} \myphi - \CArow{k} \vec{x} \right) \Big)
\end{align}
which is equivalent to solving $\argmax_{\vec{x}} E_C(\vec{x})$ where
\begin{equation} \label{eq:lcd_couplings_simplfied}
    E_C(\vec{x}) = \frac1{\ncouplings} \sum_{k=1}^{\ncouplings} \cos \left( \Crow{k} \myphi - \CArow{k} \vec{x}  \right).
\end{equation}

This formulation is equivalent to considering a composite likelihood estimate (CLE) instead of the MLE \cite{lindsay1988, varin2008, varin2011}. Composite likelihood formulas consider the likelihood of composite events (events involving a subset of the samples). Then, assuming that these composite events are independent, we can compute the likelihood of all our data as the product of these composite likelihoods.

Asymptotically, the composite likelihood can yield consistent and unique estimators under certain conditions. In particular, we consider maximizing a CLE using the pairwise sums or differences between phases \cite{lele2002}. Previous work has shown that the information from pairwise comparisons is maximized by a subset of these comparisons \cite{varin2008}. This result suggests that we do not need to consider all pairs of phases.


The distribution of the sum of random variables, each of which follows a von Mises distribution, can be approximated as a von Mises distribution centred around the sum of the means \cite{markovic2012}. Due to the symmetry of the von Mises distribution, the random variables $\Phi_{i_k}$ and $\Ik \Phi_{j_k}$ can be viewed as Von Mises random variables with mean $A_{i_k} \vec{x}$ and $\Ik A_{j_k} \vec{x}$ respectively. The variances are the same. That is, $\Phi_{i_k}  + \Ik \Phi_{j_k}$ is a von Mises random variable with mean $(A_{i_k} + \Ik A_{j_k})\vec{x}.$ The variance of $\Phi_{i_k} + \Ik \Phi_{j_k}$ changes with the amount of noise in the SSP, but is the same for all pairwise comparisons. Then, by a similar derivation to the MLE, the CLE of $\vec{x}$ given the observed phases $\myphi$ satisfies \eqref{eq:lcd_couplings_simplfied}.

In addition to this statistical interpretation, maximizing \eqref{eq:lcd_couplings_simplfied} is equivalent to maximizing the real component of $\cip{e^{iC\myphi}}{e^{i CA \vec{x}}}$, which is the similarity between the $n_c$-dimensional FHRR vectors corresponding to the observed phases $C\myphi$ and ideal phases $CA\vec{x}$. That is, we are maximizing
\begin{equation}
    S'(\vec{x}) = \mathrm{Re} \left( e^{iC\myphi} \cdot \overline{e^{i CA\vec{x}}} \right) = \mathrm{Re} \left( e^{i C(\myphi - A \vec{x})} \right) . \label{eq:real_sim_coupled}
\end{equation}

The entries of $CA$ can be interpreted as new base phases. By carefully choosing the phase couplings, this approach can result in a smoother objective function.


\subsection{Selecting Phase Couplings}





How should we couple phases to obtain a smoother objective function without sacrificing accuracy? We can control $C$ by selecting which of the approximately $n^2$ couplings we want to include in our objective function. Our selection of couplings will influence the distribution of phase differences $CA\vec{x}$ and $C\myphi$, and this distribution affects the shape of the similarity function that we will try to optimize \cite{frady2021, frady2022}.

\begin{figure}
    \centering
    \includegraphics[width=0.48\textwidth]{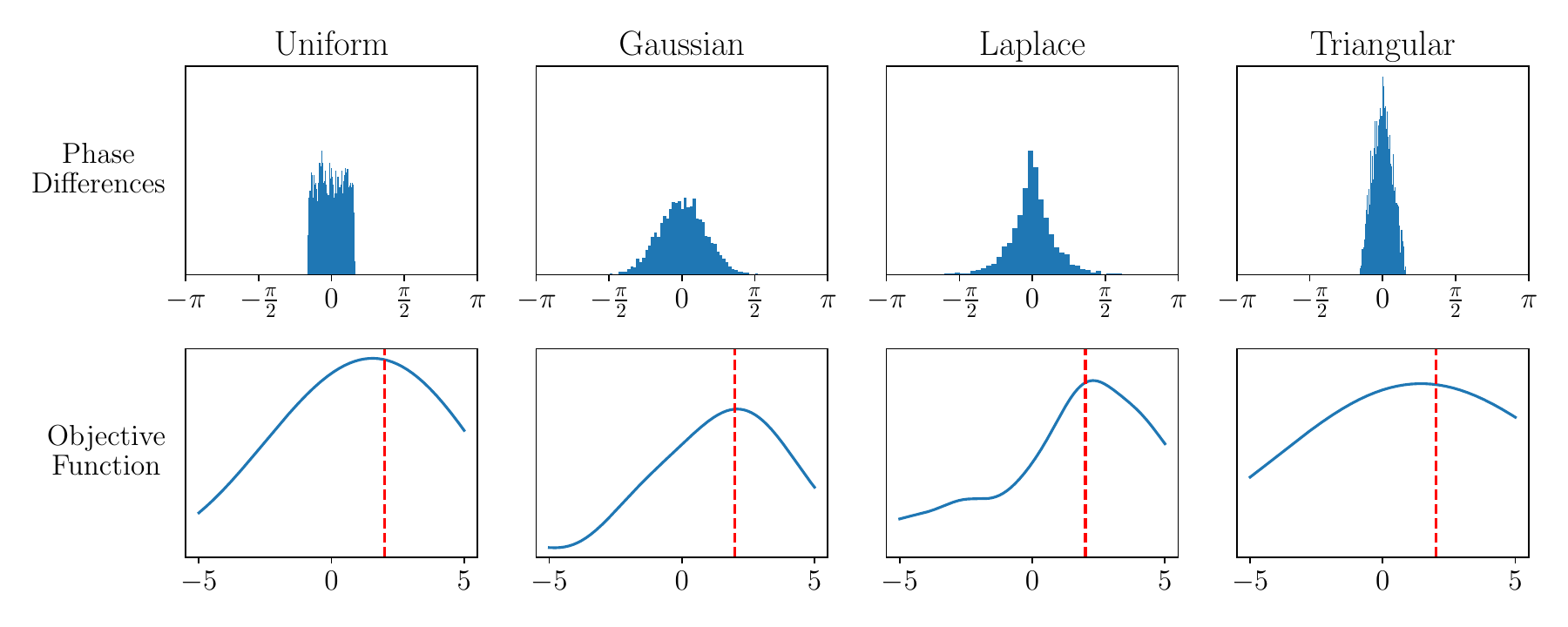}
    \caption{Histograms of the base phase combinations, and the corresponding similarity functions. The distributions plotted are the uniform, Gaussian, Laplace, and triangular. In the plots of the objective function, the red line is at $x=2$ (the true $x$ value). The SSPs were $1000$-dimensional.}
    \label{figure:phase_difference_distributions}
\end{figure}

The similarity is essentially an interference pattern between a collection of oscillating functions (where the frequency of oscillations is determined by $CA$). We can turn to Fourier theory to see that different distributions result in different similarity functions. For example, distributions that contain a lot of large phase differences will contain more high-frequency components, and generate an oscillating similarity function with many local optima. Thus, it is preferable to choose couplings with relatively small phase differences. We can even select phase differences to achieve a desired distribution of phases. Some examples (shown in Fig.\,\ref{figure:phase_difference_distributions}) include a narrow uniform distribution (or band-pass filter), a narrow triangular distribution, a scaled Gaussian distribution, or a scaled Laplace distribution. In all these cases, the phase differences are relatively close to $0$.


However, selecting only couplings with small phase differences causes problems. If all $CA$ are small, then the similarity function will be composed of only low-frequency components, and yield almost flat lines over the range of $\vec{x}$ we are considering. Then the problem is ill-conditioned, since small changes in noise can significantly change the location of the peak of the objective function.


\begin{figure}
    \centering
    \includegraphics[width=0.48\textwidth]{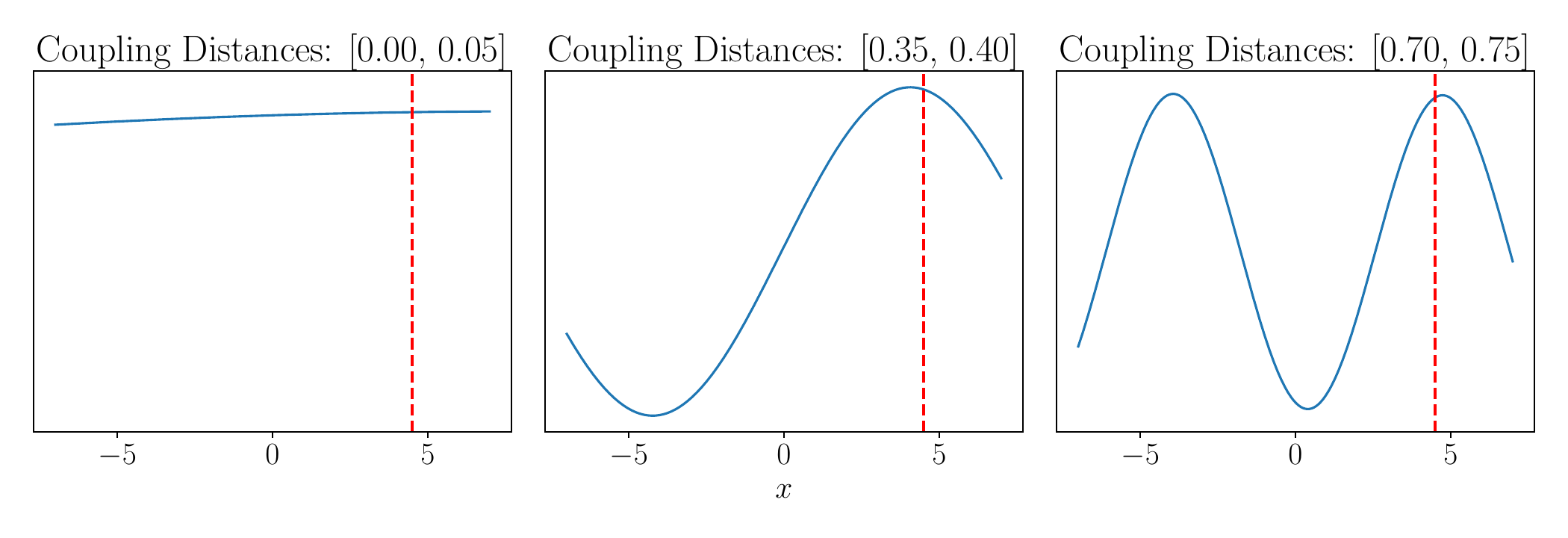}
    \caption{The objective function from coupling phases $i,j$ where the absolute value of the base phase combination $\CArow{j}$ is in the range (from left to right) [0,0.05], [0.35, 0.4] and [0.7, 0.75]. The red line is at $x=4.5$ (the true $x$ value). The SSPs were $1000$-dimensional.}
    \label{figure:coupling_distance}
\end{figure}

Figure \ref{figure:coupling_distance} demonstrates the impact of coupling distance on the objective function. Here, we are considering a $1000$-dimensional SSP and a $1$-dimensional $x$-value from $-5$ to $5$. The noisy SSP was extracted from a normalized bundle of $10$ items. The figure plots the objective function for coupled base phases with absolute value between $0$ and $0.05$, between $0.35$ and $0.40$, and between $0.70$ and $0.75$.

Figure \ref{figure:phase_difference_distributions} illustrates how the underlying distribution of the coupled phase differences affects the similarity function. To generate the desired distribution, we sample that distribution to generate a target phase difference, and then seek a new phase coupling that best matches that phase difference. 
We ensure that each base phase is involved in a prescribed minimum number of couplings. 

The bottom row in Fig.\,\ref{figure:phase_difference_distributions} plots the coupled similarity, $E_C(\vec{x})$, of a noisy SSP for a variety of different phase-difference distributions. The noisy SSP was extracted from a normalized bundle of $10$ items.

\begin{figure}
    \centering
    \includegraphics[width=0.48\textwidth]{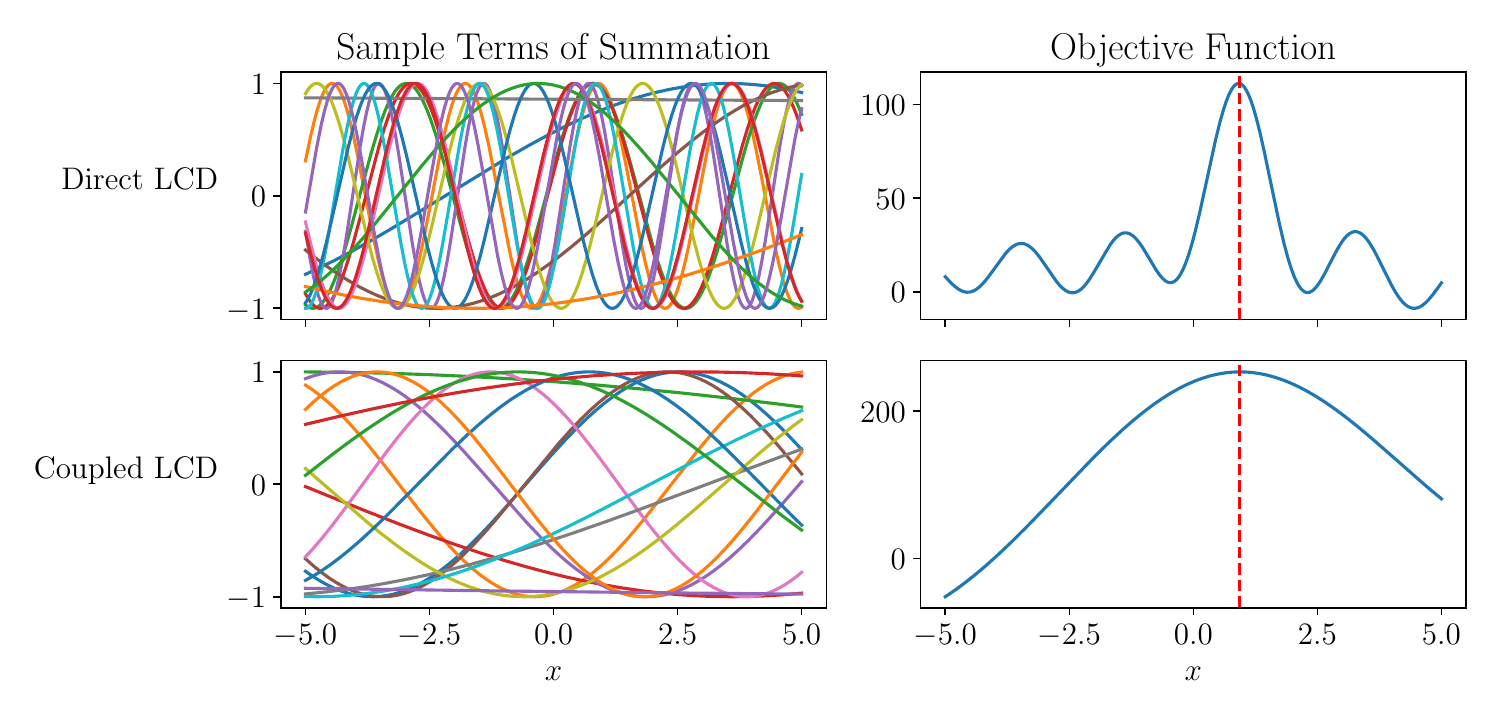}
    \caption{Left: 15 random terms from the direct least circular distance objective function (top), and 15 random terms from the coupling least circular distance objective function (bottom). Right: The two objective functions, with the true $\vec{x}$-value plotted as a red line.}
    \label{figure:objective_function_smoothness}
\end{figure}

Figure \ref{figure:objective_function_smoothness} displays 15 terms from the direct LCD formulation in equation \eqref{eq:lcd_simplfied}, along with 15 terms from the phase-coupled LCD formulation from equation \eqref{eq:lcd_couplings_simplfied}. The figure also plots the corresponding objective functions. The phase couplings were chosen to achieve a uniform distribution of base phase differences for a $1$-dimensional $x$-value in the range $[-5,5]$. The red line indicates the true $x$-value. In the direct LCD formulation, most initial $\vec{x}$ values would start in the wrong basin of attraction. The basin of attraction is much wider for the coupled formulation, although the maximum of the objective function is not precisely at the true value.

We can take advantage of both objectives functions by starting with the coupled formulation to direct us into the correct basin of attraction, and then use the direct formulation to find a more precise $\vec{x}$ value.

\subsection{Iterative Optimization}

Combining the direct formulation $E_D(\vec{x})$ from \eqref{eq:lcd_simplfied} with the phase-coupled formulation $E_C(\vec{x})$ from \eqref{eq:lcd_couplings_simplfied}, we arrive at the combined objective function
\begin{equation}
    \argmax_{\vec{x}} E(\vec{x}) \quad \mathrm{where} \quad E(\vec{x}) = \lambda E_{\text{D}}(\vec{x}) + (1-\lambda) E_{\text{C}} (\vec{x})
    \label{eq:final_objective}
\end{equation}
and $\lambda$ controls the relative weight between the direct and coupled objective functions. Setting $\lambda = 1$ yields the direct objective function, while setting $\lambda = 0$ yields the phase-coupled objective function.


We can find $\vec{x}$ using iterative optimization algorithms, such as gradient ascent or modified gradient ascent. Momentum-based updates, for example, can sometimes improve the convergence of our algorithm. We could also use Newton's method to search for critical points (where the gradient is zero). However, Newton's method would require us to invert the $d \times d$ Hessian matrix of $E(\vec{x})$ at every step, which could be difficult to implement in a neural network. Thus, to avoid matrix inversion, we focus on gradient ascent and its variants.

We begin optimizing the objective function with $\lambda$ set to 0; that is, we focus on the coupled LCD regression. After a fixed number of iterations, we set $\lambda = 1$ to perform the direct LCD regression. With the formulation above, the transition of $\lambda$ from $0$ to $1$ can be gradual, but we chose a jump discontinuity.

The gradient of $E$ with respect to $\vec{x}$ is
\begin{equation}
    \grad{\vec{x}} E = \lambda \grad{\vec{x}} E_D + (1-\lambda) \grad{\vec{x}} E_C
\end{equation}
where
\begin{align}
    \grad{\vec{x}} E_D &=  \sum_{k=1}^n \Arow{k}\trans \sin \left(  \phi_k - \Arow{k} \vec{x} \right) \label{eq:Edirect} \\
    \grad{\vec{x}} E_C &=  \sum_{j=1}^{n_c} \left( \CArow{j} \right)\trans \sin \left( \Crow{j} \myphi - \CArow{j} \vec{x} \right) \ .
\end{align}
In each iteration of gradient ascent, we can update $\vec{x}$ in the direction of the gradient, 
\begin{align}
    \vec{x} \gets \vec{x} + \gamma \grad{\vec{x}} E (\vec{x}) \ ,
\end{align}
where $\gamma$ is the step size.


Instead of decoding $\vec{x}$ explicitly, we can consider performing clean-up alone by incrementing the SSP phases. Letting $\mytheta$ be the cleaned-up phases, we rewrite our objective functions in terms of $\mytheta$ as
\begin{align}
E_D(\mytheta) &= \frac1n \sum_{i=1}^n  \cos(\phi_i - \theta_i ) \\
E_C(\mytheta) &= \frac1{\ncouplings} \sum_{k=1}^{\ncouplings} \cos \left( \Crow{k} \myphi - C_k \mytheta  \right) \ .
\end{align}
We can still use our gradients to update $\mytheta$ using
\begin{equation} \mytheta \gets \mytheta + \gamma A \grad{\vec{x}} E (\vec{x}) \end{equation}
where $\gamma$ is again the step size. Then the cleaned-up SSP is $e^{i\mytheta}$.

These gradient ascent updates can also be performed directly on complex-valued FHRR vectors -- the SSPs themselves. Suppose $\corrupt$ is the complex vector for the observed (corrupted) SSP, and $\clean$ is our cleaned-up estimate, so that $\clean = e^{iA\vec{x}}$. If we define $\tilde{E}_D$ as our objective using similarity between complex vectors, then
\begin{equation}
\tilde{E}_D (\vec{x}) = \left< \corrupt, \clean \right> = \frac{1}{n} \sum_{k=1}^n e^{i\left( \phi_k - A_k \vec{x} \right)} \ ,
\end{equation}
where the real part of $\tilde{E}_D(\vec{x})$ is equivalent to $E_D(\vec{x})$ in (\ref{eq:lcd_simplfied}). Then the gradient can be computed in the frequency domain, using
\begin{equation}
\grad{\vec{x}} \tilde{E}_D = - \frac{1}{n} \sum_{k=1}^n i \left( \Arow{k} \right)\trans \corrupt_k \overline{\clean_k} = - \frac{i}{n} A\trans \left( \corrupt \odot \overline{\clean} \right)
\end{equation}
where $\odot$ is the Hadamard product. We can similarly compute $\tilde{E}_C$. The update to $\vec{x}$ would then be based on the real part of those gradients, which is equivalent to the formula given in (\ref{eq:Edirect}). If the update to $\vec{x}$ is $\Delta \vec{x}$, then $\clean$ can be updated using $\clean \gets \clean \odot e^{iA\Delta \vec{x}}$.

\section{Experiments}\label{section:experiments}

We numerically explored and tested the methods discussed above, demonstrating the effectiveness of our method for cleaning up noisy SSPs. For each experiment, we ran $100$ trials, unless otherwise mentioned. We tested with $1024$-dimensional SSPs unless otherwise mentioned. The values encoded ($\vec{x}$) were sampled uniformly from a hypersphere with radius $5$.

In our experiments, we selected couplings of phases based on the uniform distribution, limiting the scale (width) of the distribution so that each term in the summation of \eqref{eq:lcd_couplings_simplfied} has only one optimum, as illustrated in Fig.\,\ref{figure:objective_function_smoothness}. We selected phases uniformly from the interval $\left[ \frac{-\pi}{2\sqrt{d} B}, \frac{\pi}{2\sqrt{d} B} \right]$
where $d$ is the dimension of $\vec{x}$, and $B$ is a bound on the absolute value of each entry of $\vec{x}$. We coupled each phase to $10$ other phases.


For our coupled similarity method, which we will denote \emph{CSim}, we performed gradient ascent with the phase-coupled objective function for at most $1000$ iterations, and then gradient ascent with the direct objective function for at most $1000$ iterations.





\subsection{Comparing to Other Methods}

In this section, we compare our optimization method to other candidate cleanup methods: a denoising autoencoder, a resonator network, and a grid search. We corrupted SSPs with Gaussian noise (we demonstrate our method on bundling noise in Section \ref{section:bundle_cleanup}).

We compared CSim to denoising autoencoders. To feed the FHRR vector into the denoising autoencoder, we separated the real and imaginary components, so that there were $2d$ separate inputs to the neural network. The single hidden layer had the same number of neurons, $2d$. Each training sample was created by choosing an $\vec{x}$, encoding it as an SSP, and then adding noise to the SSP. The corresponding target was either $\vec{x}$ (for decoding), or the clean SSP for $\vec{x}$ (for clean-up). The network was trained with backpropagation using the Adam optimization method \cite{Kingma2014AdamAM}. In our experiments, we trained a new network for each level of noise, although one network could be trained and tested on all noise levels.

We also compared our method to a resonator network \cite{frady2020, kent2020}. For each iteration of the resonator network, it used its current guess for $\vec{x}$ to decompose the SSP into its individual dimensions, tried to clean up each component, and then bound the components back together to reconstitute a cleaned-up version of the original SSP.
We used the variant with ordinary least squares weights.

Finally, we also performed a grid search over possible $\vec{x}$ values to find the value that maximized the similarity between the given SSP and the clean $e^{i A \vec{x}}$. We generated clean SSPs for each sample in a $d$-dimensional grid with grid spacing of $0.1$. This method can be prohibitively resource intensive, especially as the sampling density and dimension of $\vec{x}$ increases.

\begin{figure}
    \includegraphics[width=0.48\textwidth]{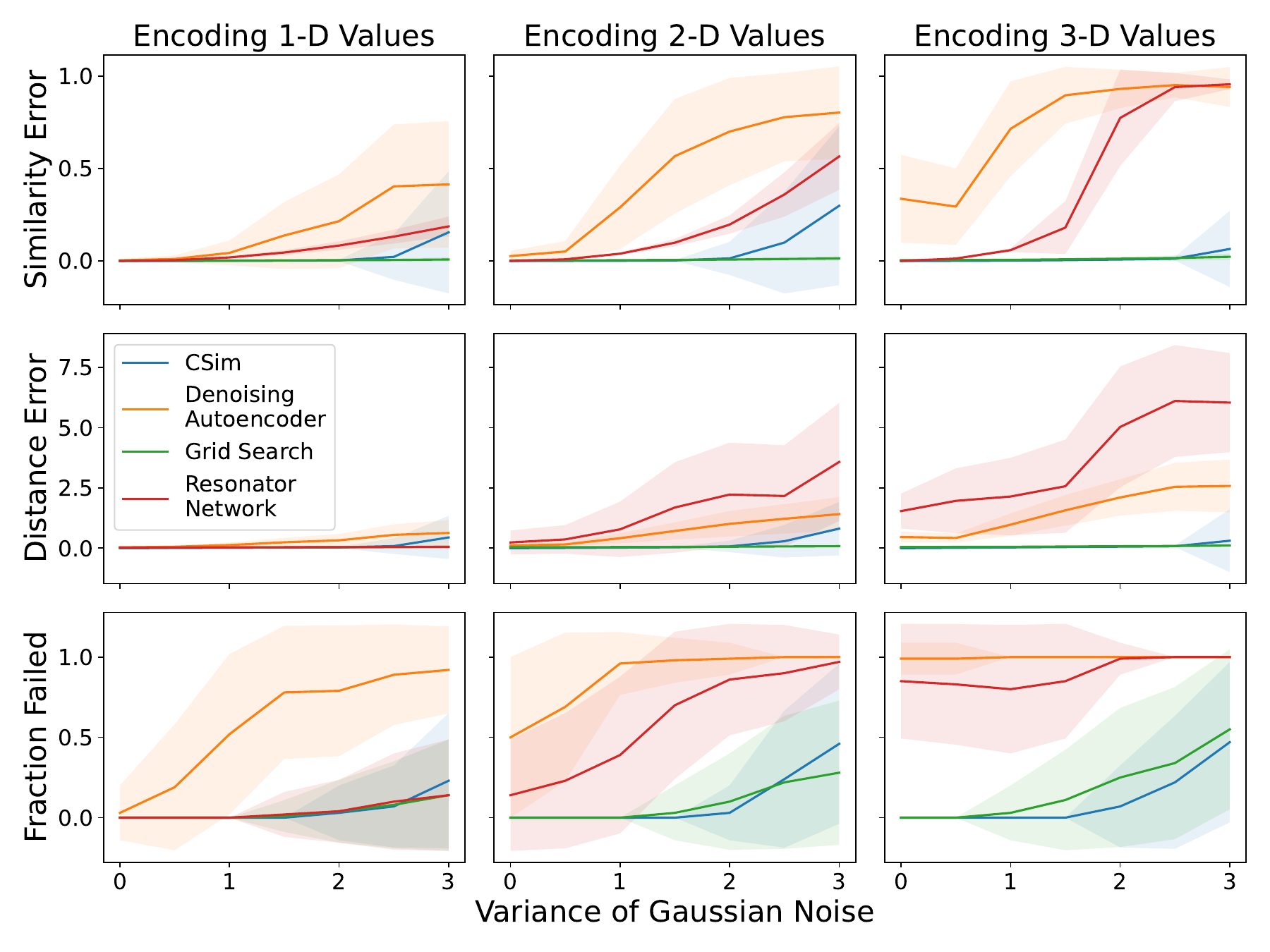}
    \caption{Similarity error, distance error, and failure fraction over a range of noise levels, averaged over 100 trials. The columns correspond to encoding 1D, 2D, and 3D values.}
    \label{figure:method_comparison}
\end{figure}

Figure \ref{figure:method_comparison} plots the performance of our method compared to other methods over a range of noise levels, averaged over 100 trials. The top row plots the similarity error, $1 - \left< \corrupt , e^{iA\vec{x}} \right>$, where $\corrupt$ is the corrupted FHRR vector and $\vec{x}$ is the decoded value.

The figure also plots the Euclidean distance between the true $\vec{x}$ value and the decoded estimate.

The bottom row of the figure shows the fraction of trials in which the decoded estimate was off by more than $0.1$. We denote such trials as \emph{fails}.
Since the resonator network does not directly return an estimate for $\vec{x}$, but instead returns SSPs for each dimension, we decoded each of these SSPs using a grid search with samples spaced $0.1$ apart, and use those value instead.

Figure \ref{figure:method_comparison} shows that our CSim method had lower similarity error than the other methods, except for grid search (which was expected because grid search performs an exhaustive -- and expensive -- search for the optimum). In terms of fraction of trials that failed, our method performed comparably to grid search and better than the other methods. Since our CSim method maximizes similarity, we expected it to converge only if the grid search method converged and the peak of the similarity function was at the true $\vec{x}$ value.

\subsection{Convergence Rate}

The gradient ascent optimization of CSim can converge quickly with the appropriate choice of step size. The step size depends on the choice of couplings, the dimension of the value encoded, and SSP dimension.

\begin{figure}
    \centering
    \includegraphics[width=0.3\textwidth]{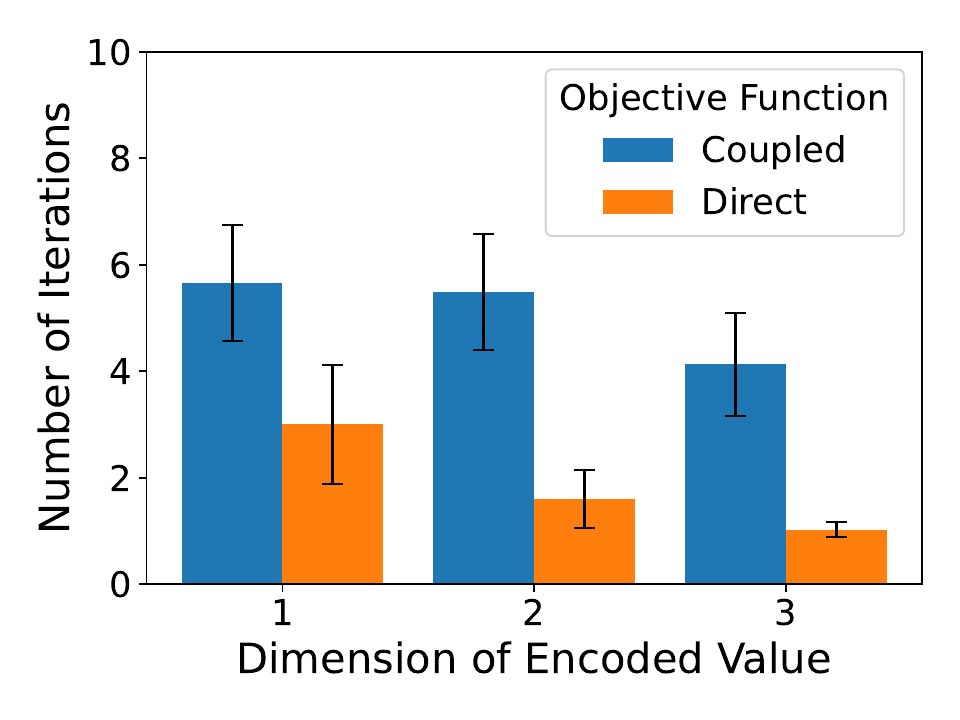}
    \caption{Number of gradient-ascent iterations to convergence.}
    \label{figure:iteration_plots}
\end{figure}

Figure \ref{figure:iteration_plots} shows the convergence speed of the two-part optimization process, averaged over 100 trials. It plots the average (and standard deviation) of the number of gradient-ascent steps of the phase-coupled objective, and the direct similarity objective, each one terminating when increments to all the elements of $\vec{x}$ are smaller than 0.01. The SSPs were corrupted with Gaussian noise with standard deviation $0.5$. 

\subsection{Cleanup of Bundle Corruption}\label{section:bundle_cleanup}

Consider constructing a bundle, $\M$, of SSPs by binding them in pairs, and then adding those together. Suppose one of those pairs is $\clean \otimes {\bf w}$. Then one could `pull out' $\clean$ by querying the bundle, unbinding it from ${\bf w}$ using $\M \oslash {\bf w}$. The result should be a noisy version of $\clean$.

Noise arises from normalizing a bundle to ensure that the resulting vector is unitary. When such a bundle is queried, the resulting vector can contain a substantial amount of noise. There is also interference noise caused by the fact that the SSPs are only approximately orthogonal.

Normalization can be done after each item is added to a bundle; this process tends to dilute the SSPs that were added early on. Alternatively, a single normalization operation done at the end of the bundling process can reduce the corruption, or at least spread it evenly over all the SSPs added to the bundle.

\begin{figure}
    \includegraphics[width=0.48\textwidth]{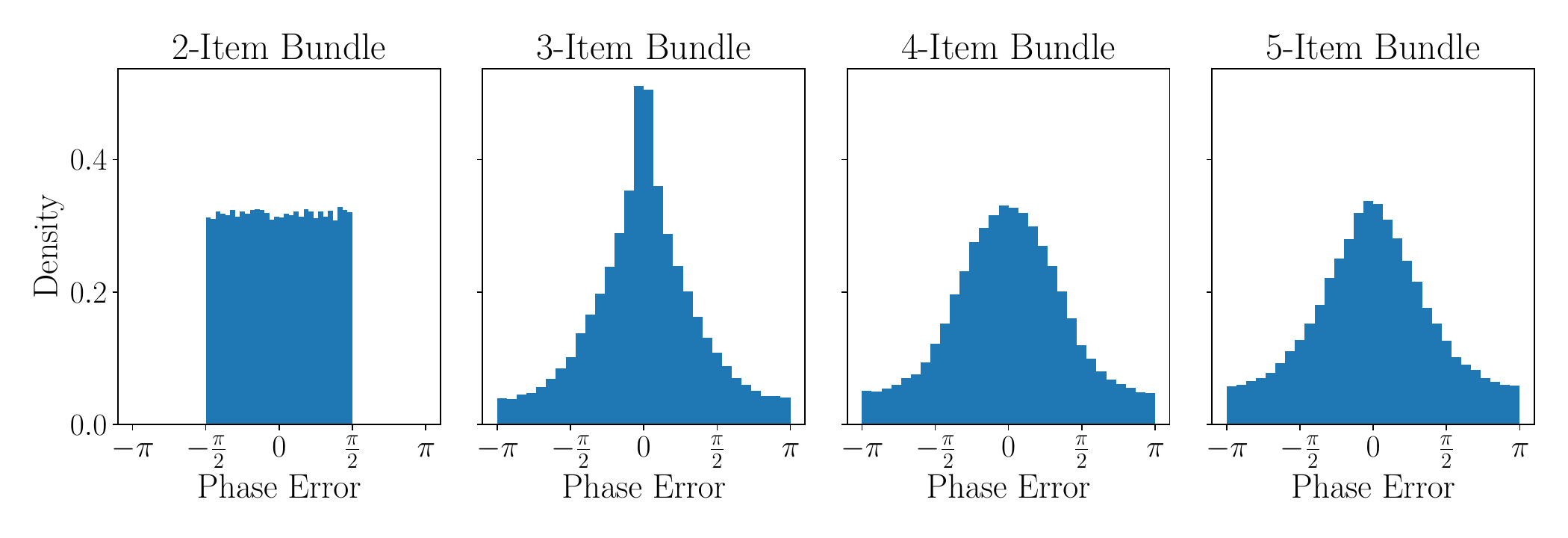}
    \caption{Histogram of phase errors for SSPs removed from bundles with $2$, $3$, $4$ and $5$ vectors (combining 100 trials of each). The phase errors have been wrapped to lie in $[-\pi,\pi]$.}
    \label{figure:phase_error}
\end{figure}

Figure \ref{figure:phase_error} displays the phase error caused by querying bundles of different numbers of items. Each bundle was normalized after adding all the items. The distribution of the phase error spreads with the number of items in the bundle.

\begin{figure}
\centering
    \includegraphics[width=0.3\textwidth]{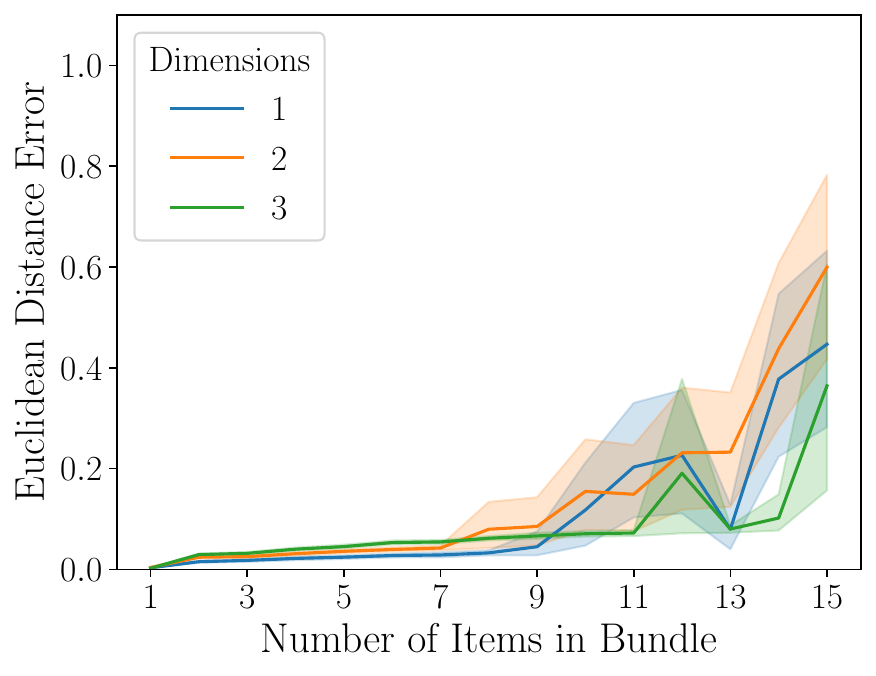}
    \caption{Euclidean distance error when cleaning up SSPs from a bundle, for different counts of bundled items. The solid line shows the mean, and the shaded region indicates 1 standard deviation.}
    \label{figure:cleanup_bundle}
\end{figure}

Bundle corruption in an SSP can affect the accuracy of the decoded value. Figure \ref{figure:cleanup_bundle} plots the average and standard deviation of the error in $\vec{x}$ (over 100 trials) when decoding from corrupted SSPs using CSim. Our method remains effective, even with increased bundling noise. 


\subsection{Demonstration of Clean-Up}

\def\p{{\bf p}}
\def\pA{\p_\symbolA}
\def\pB{\p_\symbolB}
\def\pAB{\p_{\symbolB \rightarrow \symbolA}}

\begin{figure}
    \centering
    \includegraphics[width=0.2\textwidth]{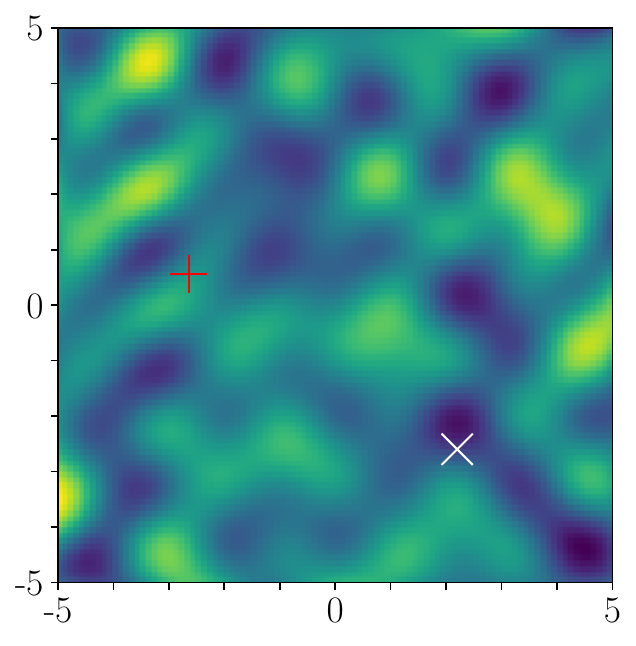}
    \includegraphics[width=0.2\textwidth]{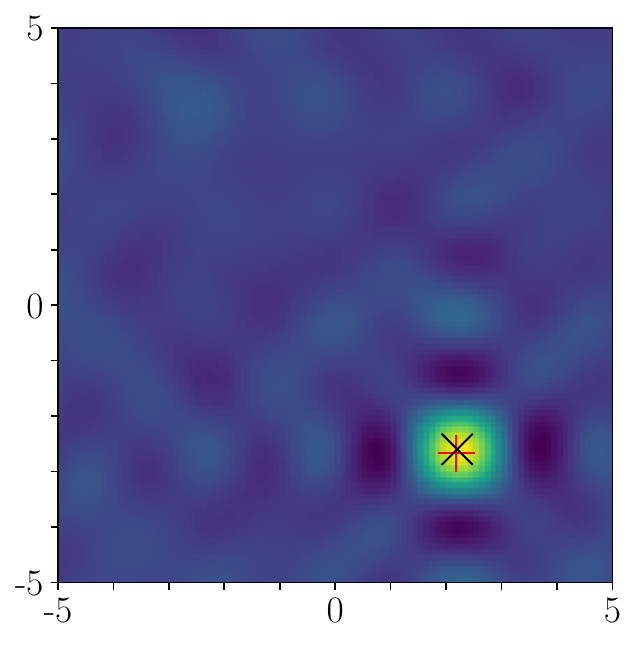}
    \caption{Similarity maps for $\pAB$ (left) and $\pAB^*$ (right). The $\times$ symbols show the true displacement, and the red $+$ symbols show the decoded displacements.}
    \label{figure:displacement}
\end{figure}

As a final test, we query the bundle
\begin{equation}
    \M =  \symbolA \otimes \X^{1.5} \otimes \Y^{-2.3} + \symbolB \otimes \X^{-0.7} \otimes \Y^{0.3}
\end{equation}
representing a \objectA \ and a \objectB \ at their corresponding $(x,y)$ positions. Suppose we want to know the displacement \emph{between} the two objects. To compute that, we query $\M$ to get the SSPs for the positions of the \objectA \ and the \objectB,
$$
\pA = \M \oslash \symbolA \quad , \quad 
\pB = \M \oslash \symbolB
$$
and then get the SSP for the displacement between them using $\pAB = \pA \oslash \pB$, which should be an encoding of $\vec{x} = (2.2, -2.6)$. However, $\pA$ and $\pB$ will contain noise that carries through to $\pAB$, resulting in a corrupted similarity map as show in Fig.\,\ref{figure:displacement} (left).

If $\pA$ and $\pB$ are first cleaned up  using CSim, giving $\pA^*$ and $\pB^*$, then $\pAB^* = \pA^* \oslash \pB^*$ yields a much better similarity map, as shown in Fig.\,\ref{figure:displacement} (right). As a result, the decoding of $\pAB^*$ using CSim correctly converges to a solution very close to $\vec{x}$. This example shows that cleaning up SSPs, as an intermediate step, results in more accurate VSA processing.

\section{Discussion and Conclusion}

These experiments demonstrate the effectiveness of our method at decoding and cleaning up SSPs. Coupling phases can result in a smoother objective function, minimizing the number of failures from starting in the wrong basin of attraction. While previous works have looked at other optimization methods \cite{kent2020}, these methods often converged to the wrong value as a result of starting in the wrong basin of attraction.

Our method of gradient ascent is efficient and easy to implement. Even with noise, our method often converges in $10$ or fewer iterations.

One main advantage of our method is that it does not need to be trained, so long as the original base phases $A$ are known. For the denoising autoencoder, training can be time-consuming and may not generalize to different levels of noise.

Our method is also guaranteed to return a true SSP regardless of $\vec{x}$ value. For the other methods that maximize similarity, they are either limited to discrete values in $\vec{x}$, as in grid search, or take linear combinations of SSPs corresponding to discrete values, as in resonator networks.

For our method to be effective, it is important to appropriately choose the phase couplings. As illustrated in Figs.\,\ref{figure:coupling_distance} and \ref{figure:phase_difference_distributions}, the distribution of their phase differences impacts the robustness of the CSim method, since it can determine the uniqueness of the peak in $E_C(\vec{x})$, and the width of its basin of attraction. We found that coupling each phase to $10$ other phases sufficiently minimized the variance, however, our method was still effective with fewer couplings.


We used a uniform phase-difference distribution for our experiments. But other distributions might perform better. For example, if we choose our phase differences so that their distribution follows a $\mathrm{sinc}^2$ function, the resulting similarity function is approximately triangular and may be easy to optimize \cite{frady2021}.

One could also explore other optimization methods. Due to the challenges with matrix inversion, we did not further investigate using Newton's method to solve this optimization problem. However, Newton's method, other variants of gradient ascent, or more sophisticated iterative optimization methods might be more efficient or effective.

We also plan to explore possible neural implementations of this method. In particular, this gradient ascent method, which performs addition directly on phases, seems particularly well-suited for computation with neurons that represent complex numbers by their phase \cite{frady2019, orchard2023hd,Orchard2024}.

Finally, while this method was initially inspired by cells in the hippocampus and proposed methods for path integration, the final method does not clearly map onto a neural system. Future work may explore if this method is more closely linked to a biological neural system.

Overall, our decoding and cleanup method efficiently and accurately finds the optimal $\vec{x}$ (or its clean SSP counterpart) better than alternative methods. The potential for neural implementation positions this method well for use in neuromorphic systems.
\bibliographystyle{IEEEtran}
\bibliography{refs}

\begin{thebibliography}{10}
\providecommand{\url}[1]{#1}
\csname url@samestyle\endcsname
\providecommand{\newblock}{\relax}
\providecommand{\bibinfo}[2]{#2}
\providecommand{\BIBentrySTDinterwordspacing}{\spaceskip=0pt\relax}
\providecommand{\BIBentryALTinterwordstretchfactor}{4}
\providecommand{\BIBentryALTinterwordspacing}{\spaceskip=\fontdimen2\font plus
\BIBentryALTinterwordstretchfactor\fontdimen3\font minus
  \fontdimen4\font\relax}
\providecommand{\BIBforeignlanguage}[2]{{%
\expandafter\ifx\csname l@#1\endcsname\relax
\typeout{** WARNING: IEEEtran.bst: No hyphenation pattern has been}%
\typeout{** loaded for the language `#1'. Using the pattern for}%
\typeout{** the default language instead.}%
\else
\language=\csname l@#1\endcsname
\fi
#2}}
\providecommand{\BIBdecl}{\relax}
\BIBdecl

\bibitem{smolensky1990}
P.~Smolensky, ``Tensor product variable binding and the representation of
  symbolic structures in connectionist systems,'' \emph{Artificial
  Intelligence}, vol.~46, 1990.

\bibitem{kanerva2009}
P.~Kanerva, ``Hyperdimensional computing: An introduction to computing in
  distributed representation with high-dimensional random vectors,''
  \emph{Cognitive Computation}, vol.~1, 2009.

\bibitem{eliasmith2013}
C.~Eliasmith, \emph{How to build a brain: A neural architecture for biological
  cognition}.\hskip 1em plus 0.5em minus 0.4em\relax New York, NY: Oxford
  University Press, 2013.

\bibitem{Gayler2003}
R.~W. Gayler, ``Vector symbolic architectures answer jackendoff's challenges
  for cognitive neuroscience,'' in \emph{Joint International Conference on
  Cognitive Science}, 2003, pp. 133--138.

\bibitem{Komer2019}
B.~Komer, T.~C. Stewart, and C.~Eliasmith, ``A neural representation of
  continuous space using fractional binding,'' in \emph{Proceedings of the 41st
  Annual Meeting of the Cognitive Science Society}, 2019.

\bibitem{lu2019}
T.~Lu, A.~R. Voelker, B.~Komer, and C.~Eliasmith, ``Representing spatial
  relations with fractional binding,'' in \emph{Proceedings of the Annual
  Meeting of the Cognitive Science Society}, 2019.

\bibitem{komer2020}
B.~Komer and C.~Eliasmith, ``Efficient navigation using a scalable,
  biologically inspired spatial representation,'' in \emph{42nd Annual Meeting
  of the Cognitive Science Society}.\hskip 1em plus 0.5em minus 0.4em\relax
  Cognitive Science Society, 2020.

\bibitem{voelker2021a}
A.~R. Voelker, P.~Blouw, X.~Choo, N.~S.-Y. Dumont, T.~C. Stewart, and
  C.~Eliasmith, ``Simulating and predicting dynamical systems with spatial
  semantic pointers,'' \emph{Neural Computation}, vol.~33, no.~8, pp.
  2033--2067, 07 2021.

\bibitem{Stewart2011}
T.~C. Stewart, Y.~Tang, and C.~Eliasmith, ``A biologically realistic cleanup
  memory: Autoassociation in spiking neurons,'' \emph{Cognitive Systems
  Research}, vol.~12, pp. 84--92, 2011.

\bibitem{frady2020}
E.~P. Frady, S.~J. Kent, B.~A. Olshausen, and F.~T. Sommer, ``Resonator
  networks, 1: An efficient solution for factoring high-dimensional,
  distributed representations of data structures,'' \emph{Neural Computation},
  vol.~32, no.~12, pp. 2311--2331, 2020.

\bibitem{kleyko2022}
D.~Kleyko, M.~Davies, E.~P. Frady, P.~Kanerva, S.~J. Kent, B.~A. Olshausen,
  E.~Osipov, J.~M. Rabaey, D.~A. Rachkovskij, A.~Rahimi, and F.~T. Sommer,
  ``Vector symbolic architectures as a computing framework for emerging
  hardware,'' \emph{Proceedings of the IEEE}, vol. 110, no.~10, pp. 1538--1571,
  2022.

\bibitem{kent2020}
S.~J. Kent, E.~P. Frady, F.~T. Sommer, and B.~A. Olshausen, ``{Resonator
  Networks, 2: Factorization Performance and Capacity Compared to
  Optimization-Based Methods},'' \emph{Neural Computation}, vol.~32, no.~12,
  pp. 2332--2388, 2020.

\bibitem{lund1999}
U.~Lund, ``Least circular distance regression for directional data,''
  \emph{Journal of Applied Statistics}, vol.~26, no.~6, pp. 723--733, 1999.

\bibitem{burgess2008}
N.~Burgess, ``Grid cells and theta as oscillatory interference: Theory and
  predictions,'' \emph{Hippocampus}, vol.~18, no.~12, pp. 1157--1174, 2008.

\bibitem{OrchardFrontCompNeuro13}
J.~Orchard, H.~Yang, and X.~Ji, ``Does the entorhinal cortex use the {Fourier}
  transform?'' \emph{Frontiers in computational neuroscience}, vol.~7, 12 2013.

\bibitem{burgess2014}
\BIBentryALTinterwordspacing
C.~P. Burgess and N.~Burgess, ``Controlling phase noise in oscillatory
  interference models of grid cell firing,'' \emph{Journal of Neuroscience},
  vol.~34, no.~18, pp. 6224--6232, 2014. [Online]. Available:
  \url{https://www.jneurosci.org/content/34/18/6224}
\BIBentrySTDinterwordspacing

\bibitem{Ji2014}
X.~Ji, ``Generalized strategies for path integration using neural
  oscillators,'' Master's thesis, University of Waterloo, 2014.

\bibitem{dumont2022model}
N.~S.-Y. Dumont, J.~Orchard, and C.~Eliasmith, ``A model of path integration
  that connects neural and symbolic representation,'' in \emph{Proceedings of
  the Annual Meeting of the Cognitive Science Society}, vol.~44, no.~44, 2022.

\bibitem{Plate1995}
T.~A. Plate, ``Holographic reduced representations,'' \emph{IEEE Transactions
  on Neural Networks}, vol.~6, pp. 623--641, 1995.

\bibitem{schlegel2022}
K.~Schlegel, P.~Neubert, and P.~Protzel, ``A comparison of vector symbolic
  architectures,'' \emph{Artificial Intelligence Review}, vol.~55, 2022.

\bibitem{plate1992}
T.~A. Plate, ``Holographic recurrent networks,'' in \emph{Advances in Neural
  Information Processing Systems}, vol.~5.\hskip 1em plus 0.5em minus
  0.4em\relax Morgan-Kaufmann, 1992.

\bibitem{plate1994}
------, ``Distributed representations and nested compositional structure,''
  Ph.D. dissertation, University of Toronto, 1994.

\bibitem{furlong2022fractional}
P.~M. Furlong and C.~Eliasmith, ``Fractional binding in vector symbolic
  architectures as quasi-probability statements,'' in \emph{44th Annual Meeting
  of the Cognitive Science Society}.\hskip 1em plus 0.5em minus 0.4em\relax
  Cognitive Science Society, 2022.

\bibitem{dumont2023b}
N.~S.-Y. Dumont, P.~M. Furlong, J.~Orchard, and C.~Eliasmith, ``Exploiting
  semantic information in a spiking neural slam system,'' in \emph{Frontiers in
  Neuroscience}, vol.~17, 2023.

\bibitem{Sreenivasan2011}
S.~Sreenivasan and I.~Fiete, ``Grid cells generate an analog error-correcting
  code for singularly precise neural computation,'' \emph{Nature Neuroscience},
  vol.~14, pp. 1330--1337, 2011.

\bibitem{lindsay1988}
B.~G. Lindsay, ``Composite likelihood methods,'' \emph{Contemporary
  Mathematics}, 1988.

\bibitem{varin2008}
C.~Varin, ``On composite marginal likelihoods,'' \emph{AStA Advances in
  Statistical Analysis}, vol.~92, pp. 1--28, 2008.

\bibitem{varin2011}
C.~Varin, N.~Reid, and D.~Firth, ``An overview of composite likelihood
  methods,'' \emph{Statistica Sinica}, vol.~21, pp. 5--42, 2011.

\bibitem{lele2002}
S.~Lele and M.~L. Taper, ``A composite likelihood approach to (co)variance
  components estimation,'' \emph{Journal of Statistical Planning and
  Inference}, vol. 103, pp. 117--135, 2002.

\bibitem{markovic2012}
I.~Marković and I.~Petrovic, ``Bearing-only tracking with a mixture of von
  mises distributions,'' in \emph{2012 IEEE/RSJ International Conference on
  Intelligent Robots and Systems}, 2012, pp. 707--712.

\bibitem{frady2021}
E.~P. Frady, D.~Kleyko, C.~J. Kymn, B.~A. Olshausen, and F.~T. Sommer,
  ``Computing on functions using randomized vector representations,''
  \emph{arXiv e-prints}, 2021.

\bibitem{frady2022}
------, ``Computing on functions using randomized vector representations (in
  brief),'' in \emph{Neuro-Inspired Computational Elements Conference}.\hskip
  1em plus 0.5em minus 0.4em\relax Association for Computing Machinery, 2022,
  p. 115–122.

\bibitem{Kingma2014AdamAM}
\BIBentryALTinterwordspacing
D.~P. Kingma and J.~Ba, ``Adam: A method for stochastic optimization,''
  \emph{CoRR}, vol. abs/1412.6980, 2014. [Online]. Available:
  \url{https://api.semanticscholar.org/CorpusID:6628106}
\BIBentrySTDinterwordspacing

\bibitem{frady2019}
E.~P. Frady and F.~T. Sommer, ``Robust computation with rhythmic spike
  patterns,'' \emph{Proceedings of the National Academy of Sciences}, vol. 116,
  no.~36, pp. 18\,050--18\,059, 2019.

\bibitem{orchard2023hd}
J.~Orchard and R.~Jarvis, ``Hyperdimensional computing with spiking-phasor
  neurons,'' in \emph{Proceedings of the 2023 International Conference on
  Neuromorphic Systems}, 2023.

\bibitem{Orchard2024}
J.~Orchard, P.~M. Furlong, and K.~Simone, ``Efficient hyperdimensional
  computing with spiking phasors,'' \emph{Neural Computation}, vol.~36, no.~9,
  pp. 1886--1911, 9 2024.

\end{thebibliography}

\end{document}